\documentclass[10pt,twocolumn,letterpaper]{article}

\usepackage{cvpr}
\usepackage{times}
\usepackage{epsfig}
\usepackage{graphicx}
\usepackage{amsmath}
\usepackage{amssymb}
\usepackage{subcaption}

\usepackage{multirow}
\usepackage{fancyhdr}

\DeclareMathOperator{\E}{\mathbb{E}}
\graphicspath{{images/}}

\usepackage[font=small,skip=1pt]{caption}
\usepackage{floatrow}
\usepackage[pagebackref=true,breaklinks=true,letterpaper=true,colorlinks,bookmarks=false]{hyperref}

\cvprfinalcopy % *** Uncomment this line for the final submission

\def\cvprPaperID{4423} % *** Enter the CVPR Paper ID here

\ifcvprfinal\fi
\begin{document}
	
	%%%%%%%%% TITLE
	\title{REPAIR: Removing Representation Bias by Dataset Resampling}
	
	\author{Yi Li\\
		UC San Diego\\
% 		Institution1 address\\
		{\tt\small yil898@ucsd.edu}
		\and
		Nuno Vasconcelos\\
		UC San Diego\\
% 		First line of institution2 address\\
		{\tt\small nvasconcelos@ucsd.edu}
	}
	
	\maketitle

	\thispagestyle{fancy} 
	\renewcommand{\headrulewidth}{0pt}
	\lfoot{\small To appear in \textit{Conference on Computer Vision and Pattern Recognition} (CVPR), Long Beach, 2019.}
	
	%%%%%%%%% ABSTRACT
	\begin{abstract}
		Modern machine learning datasets can have biases for certain representations
		that are leveraged by
		algorithms to achieve high performance without
		learning to solve the underlying task. This problem is referred to
		as ``representation bias''. The question of how to reduce the representation 
		biases of a dataset is investigated and a new dataset {\it REPresentAtion
			bIas Removal\/} (REPAIR) procedure is proposed. This
		formulates bias minimization as an optimization problem,
		seeking a weight distribution that penalizes examples 
		easy for a classifier built on a given feature representation. 
		Bias reduction is then equated to maximizing the ratio between the
		classification loss on the reweighted dataset and the uncertainty of 
		the ground-truth class labels. This is a minimax problem
		that REPAIR solves by alternatingly updating classifier parameters
		and dataset resampling weights, using stochastic gradient descent. 
		An experimental set-up is also introduced to measure
		the bias of any dataset for a given representation,
		and the impact of this bias on the performance of recognition
		models. Experiments with synthetic and action recognition
		data show that dataset REPAIR can significantly reduce representation
		bias, and lead to improved generalization of models trained on REPAIRed
		datasets. The tools used for characterizing representation bias, and the proposed dataset REPAIR algorithm, are available at {\small\url{https://github.com/JerryYLi/Dataset-REPAIR/}}.
	\end{abstract}
	
	%%%%%%%%% BODY TEXT
	\vspace{-1em}
	\section{Introduction}
	Over the last decade, deep neural networks (DNNs) have enabled 
	transformational advances in various fields, delivering superior 
	performance on large-scale benchmarks. However like any other machine 
	learning systems, the quality of DNNs is only as good as that of the 
	datasets on which they are trained. In this regard, there are at least two 
	sources of concern. First, they can have limited generalization 
	beyond their training domain \cite{torralba2011unbiased,beery2018recognition}. 
	This is classically known as {\it dataset bias\/}. Second, the learning procedure could give rise to biased deep learning 
	algorithms \cite{bolukbasi2016man,ritter2017cognitive}.
	\emph{Representation bias} is an instance of this problem, that follows from
	training on datasets that favor certain representations over 
	others \cite{li2018resound}. When a dataset is easily solved by adoption
	of a specific feature representation $\phi$, it is said to be biased 
	towards $\phi$. Bias is by itself not negative: If the classification
	of \textit{scenes}, within a certain application context, is highly 
	dependent on the detection of certain \textit{objects}, 
	successful scene recognition systems are likely to require
	detailed object representations. In this application context, scene
	recognition datasets should exhibit \textit{object bias}.
	However, in the absence of mechanisms to measure and control bias, it 
	is unclear if conclusions derived from experiments are tainted by 
	undesirable biases. When this is the case, learning algorithms could 
	simply overfit to the dataset biases, hampering generalization
	beyond the specific dataset.
	
	This problem is particularly relevant for action recognition, 
	where a wide range of diverse visual cues can be informative of action
	class labels, and leveraged by different algorithms. 
	In the literature, different algorithms tend to implement different
	representations. Some models infer action categories from one or a few
	video frames \cite{simonyan2014two,huang2018makes,zhou2018temporal}, while 
	others attempt to model long-term dependencies \cite{wang2013action,
		yue2015beyond,girdhar2017actionvlad}; some focus on modeling human 
	pose \cite{jhuang2013towards}, and some prefer to incorporate contextual 
	information \cite{gkioxari2015contextual}. In general, two algorithms 
	that perform equally well on a dataset biased towards a representation, \eg
	a dataset with {\it static\/} or single frame bias, can behave in a
	drastically different manner when the dataset is augmented 
	with examples that eliminate this bias, \eg by requiring more temporal 
	reasoning. Without the ability to control the static bias of the dataset, 
	it is impossible to rule out the possibility that good performance
	is due to the ability of algorithms to pick up 
	spurious static visual cues (\eg backgrounds, objects, \etc)
	instead of modeling action.

	In this work, we investigate the question of how to reduce the representation 
	biases of a dataset. For this, we introduce a new {\it REPresentAtion
		bIas Removal\/} (REPAIR) procedure for dataset resampling, based on an 
	a formulation of bias minimization as an optimization problem.
	REPAIR seeks a set of example-level weights penalizing examples that are 
	easy for a classifier built on a given feature representation. 
%	This is implemented by using a DNN to implement a feature
%	extractor for the representation of interest and learning an
%	independent linear classifier to classify these features. 
	This is implemented by using a DNN as feature
	extractor for the representation of interest and learning an
	independent linear classifier to classify these features. 
	Bias reduction is then equated to maximizing the ratio between the loss 
	of this classifier on the reweighted dataset and the uncertainty of 
	the ground truth class labels. We show that this reduces to a 
	minimax problem, solved by 
	alternatingly updating the classifier coefficients and the dataset 
	resampling weights, using stochastic gradient descent (SGD). 
	
	Beyond introducing the dataset REPAIR procedure, we develop an experimental
	procedure for its evaluation. We consider two scenarios in this work. The first 
	is a controlled experiment where we explicitly add \emph{color bias} 
	to an otherwise unbiased dataset of grayscale images. This enables
	the design of experiments that explicitly measure recognition 
	performance as a function of the amount of bias.
	The second is action recognition from videos, where many
	popular datasets are known to have \emph{static bias}. In both cases,
	dataset REPAIR is shown to substantially reduces representation bias,
% 	outperforming random sampling by a significant margin.
	which is not possible with random subsampling.
	A generic set-up is then introduced to evaluate the effect of 
	representation bias on model training and evaluation. 
	This has two main components. The first measures how the performance 
	of different algorithms varies as a function of the bias of datasets towards a given representation. The second analyzes how representation bias affects 
	the ability of algorithms to generalize across datasets.
	Various experiments in this set-up are performed
	leading to a series of interesting findings about behavior of models on resampled 
	datasets.
	
	Overall, the paper makes three main contributions. The first is a
	novel formulation of representation bias minimization as a differentiable 
	and directly optimizable problem. The second is a SGD-based dataset 
	resampling strategy, REPAIR, which is shown able to significantly reduce 
	representation bias. The third is a new experimental set-up for 
	evaluating dataset resampling algorithms, that helps determine
	the importance of such resampling to achieving both model generalization and 
	fair algorithm comparisons. 
	
	\section{Related Work}
	
	\paragraph{Fair Machine Leaning.}
	As data-driven learning systems are used in an increasingly
	larger array of real-world applications, the fairness and bias of the decisions
	made by these systems becomes an important topic of study. In recent years,
	different criteria have been proposed to assess the fairness of
	learning algorithms \cite{zemel2013learning,feldman2015certifying,
		hardt2016equality}, stimulating attempts to build unbiased algorithms.
	In general, deep learning systems are apt at capturing or even magnifying
	biases in their supervisory information \cite{ritter2017cognitive,
		zhao2017men,anne2018women,stock2018convnets}. This is in part due to the
	end-to-end nature of their training, which encourages models to exploit
	biased features if this leads to accurate classification. Prior works
	have mostly focused on uncovering and addressing different instances of
	bias in learned models, including gender
	bias \cite{bolukbasi2016man,zhao2017men,anne2018women} and racial
	bias \cite{stock2018convnets}. However, the bias of the data itself has
	received less attention from the community.
	
	\vspace{-1em}
	\paragraph{Dataset Bias.}
	While datasets are expected to resemble the probability distribution of
	observations, the data collection procedure can be biased by
	human and systematic factors, leading to distribution mismatch between
	dataset and reality, as well as between two datasets. This is
	referred to as dataset bias \cite{torralba2011unbiased,tommasi2017deeper}.
	\cite{torralba2011unbiased} analyzed the forms of bias
	present in different image recognition datasets, and demonstrated its
	negative effect on cross-dataset model generalization.
	Dataset bias has been well studied and can be compensated with domain
	adaptation techniques \cite{khosla2012undoing,fernando2013unsupervised,
	oquab2014learning}.
	
	Representation bias is a more recent concept, describing the ability of
	a representation to solve a dataset. It was first explicitly
	formulated in \cite{li2018resound}, and used to measure the bias of
	modern action recognition datasets towards objects, scenes and people.
% 	This work called for the minimization of representation biases, but did
% 	not produce a concrete algorithm to do this. 
	Representation bias is different from dataset bias, in that it enables potential ``shortcuts''
	(the representations for which the dataset is biased) that a model can
	exploit to solve the dataset, without learning the underlying task of
	interest. For example, contextual bias allows recognition algorithms to
	recognize objects by simply observing their environment
	\cite{torralba2003contextual}. Even when an agent does not
	rely solely on shortcuts, its decisions may be biased for these
	representations, as \cite{ritter2017cognitive} showed in their case study of
	how shape bias is captured by models trained on ImageNet. 
	
	\vspace{-1em}
	\paragraph{Video Action Recognition.}
	Early efforts at human action recognition mainly relied on compact video
	descriptors encoding hand-crafted spatiotemporal
	features \cite{laptev2005space,wang2011action,wang2013action}. Deep
	learning approaches, like two-stream networks \cite{simonyan2014two},
	3D convolutional networks \cite{ji20133d,tran2015learning} and recurrent
	neural networks \cite{yue2015beyond}, use network architectures that
	learn all relevant features. A common theme across many action recognition
	works is to capture long-term temporal structure in the video.
	However, current datasets have an abundance of static cues that can give
	away the action (\ie bias towards static representations),
	making it difficult to assess the importance of long-term temporal
	modeling. The presence of this static bias has been noted and studied
	in previous work: \cite{gkioxari2015contextual} exploited contextual cues
	to achieve state-of-the-art action recognition performance.
	\cite{feichtenhofer2018have} visualized action models to uncover unwanted
	biases in training data. Finally, \cite{huang2018makes} identified
	action categories that can be recognized without any temporal reasoning.
	
	\vspace{-1em}
	\paragraph{Dataset Resampling.}
	Resampling refers to the practice of obtaining sample points with different
	frequencies than those of the original distribution. It is commonly
	used in machine learning to balance datasets, by oversampling minority
	classes and under-sampling majority ones \cite{chawla2002smote}.
	By altering relative frequencies of examples, dataset resampling
	enables the training of fairer models, which do not discriminate
	against minority classes.
	
	\section{Minimum-bias Dataset Resampling}
	
	\subsection{Representation Bias}
	\emph{Representation bias} \cite{li2018resound} captures the bias of
	a dataset with respect to a representation. Let $\phi: \mathcal{X} \to
	\mathcal{Z}$ be a feature 
	representation. The bias of dataset $\mathcal{D}$ towards $\phi$ is the 
	\emph{best achievable performance} of the features $\phi(x)$ on
	$\mathcal{D}$ normalized by chance level. In this work, we 
	measure classification performance with the risk defined by the
	cross-entropy loss
	\begin{equation}
	\mathcal{R}^*(\mathcal{D}, \phi) = \min_\theta\ 
	\E_{X, Y}[-\log P(Y \mid Z; \theta)]
	\label{eqn:risk}
	\end{equation}
	where $X$ and $Y$ are examples and their respective labels, and $Z = \phi(X)$ is the feature-space representation of $X$. Here $P(Y \mid \phi(X); \theta)$ is computed by a softmax layer (weight matrix plus softmax nonlinearity) of input $Z$ and parameters $\theta$, which are optimized by gradient descent. 
	We do not fine-tune the representation $\phi$ itself 
	to retain its original semantics; only the parameters of the softmax layer are learned.
	Noting that minimizing the cross-entropy loss encourages the softmax classifier to output the true posterior class probabilities $P(Y \mid Z)$, we may rewrite \eqref{eqn:risk} as
	\begin{align}
	\mathcal{R}^*(\mathcal{D}, \phi) &= \E_{Z, Y}[-\log P(Y \mid Z)] \nonumber \\
	&= \E_{Z, Y}\left[-\log P(Y) - \log \frac{P(Z, Y)}{P(Z) P(Y)}\right] \nonumber \\
	&= H(Y) - I(Z, Y)
	\end{align}
	The risk $\mathcal{R}^*(\mathcal{D}, \phi)$ is therefore
	upper-bounded by the entropy of class label $Y$ and decreases as 
	the mutual information between the feature vector $Z$ and the label $Y$ 
	increases. Hence, a lower $\mathcal{R}^*(\mathcal{D}, \phi)$ indicates that
	$\phi$ is  more informative for solving $\mathcal{D}$, \ie the representation 
	bias is larger. 
	% The above holds under the assumption that the learned classifier well approximates the true distribution $P(Y \mid Z)$, which is reasonable for compact, high-level feature representations like convnet features.
	This is captured by defining bias as
	\begin{equation}
	\mathcal{B}(\mathcal{D}, \phi) = \frac{I(Z, Y)}{H(Y)} 
	= 1 - \frac{\mathcal{R}^*(\mathcal{D}, \phi)}{H(Y)}.
	\label{eqn:bias}
	\end{equation}
	Intuitively, bias has a value in $[0, 1]$ that characterizes 
	the \emph{reduction in uncertainty} about the class label $Y$ when 
	feature $Z$ is observed. The normalization term $H(Y)$ guarantees
	fairness of bias measurements when datasets have different numbers of classes. 
	In practice the terms used to define bias \eqref{eqn:bias} are estimated by their empirical values
	\begin{align}
	\mathcal{R}^*(\mathcal{D}, \phi) 
	&\approx \min_\theta 
	-\frac{1}{|\mathcal{D}|} \sum_{({\bf x}, y) \in \mathcal{D}} 
	\log P(y \mid {\bf x}; \theta) \label{eqn:emp_risk} \\
	H(Y) &\approx -\frac{1}{|\mathcal{D}|} \sum_{({\bf x}, y) \in \mathcal{D}} \log p_y \label{eqn:entropy}
	\end{align}
	where $p_y$ is the frequency of class $y$. Measuring the bias thus
	amounts to learning a linear classifier $\theta$, referred to
	as \textit{bias estimator}, and recording its cross-entropy loss as well
	as the class frequencies. It should be noted that the bias formulation of
	\eqref{eqn:bias} differs from that of \cite{li2018resound}, in that 1)
	the bias value is properly normalized to the range $[0, 1]$, and 2)
	bias is differentiable \wrt $\theta$. The last property is
	particularly important, as it enables bias optimization.
	
	\subsection{Adversarial Example Reweighting}
	Representation bias can be problematic because it implies that the
	dataset $\cal D$ favors some representations over others.
	While there is an unknown ground-truth representation $\phi^*$
	that achieves the best performance on a task, this may not be the case
	for a dataset ${\cal D}$ of that task, if the dataset is biased towards
	other representations. We provide some simple examples of this in
	Sections~\ref{sect:colored_mnist} and \ref{sect:action}. When this is the
	case, it is  desirable to modify the dataset so as to minimize bias.
	% favors a representation other  it is important to minimize the bias of a dataset towards those representations 1) irrelevant to the defining characteristics of data categories, but 2) containing cues that ``give away'' the ground-truth. Extensive experiments in Sections \ref{sect:colored_mnist} and \ref{sect:action} will further exemplify how biased datasets might prevent deep models from learning crucial features.
	One possibility, that we explore in this work, is to perform 
	dataset resampling. While the risk of \eqref{eqn:emp_risk} and entropy
	of \eqref{eqn:entropy} assign equal weight to each example
	in ${\cal D}$, bias can be controlled by prioritizing certain examples 
	over others. In other words, we attempt to create a new dataset 
	$\mathcal{D}'$ of reduced bias, by non-uniformly sampling examples
	from the existing dataset $\cal D$. For this, it suffices to augment
	each example $({\bf x}_i, y_i) \in \mathcal{D}$ with a weight $w_i$ 
	that encodes the probability of the example being selected by the resampling 
	procedure. 
	This transforms \eqref{eqn:emp_risk} and \eqref{eqn:entropy} into
	\begin{align}
	\mathcal{R}^*(\mathcal{D}', \phi) 
	&\approx \min_\theta - 
	\sum_{i=1}^{|\mathcal{D}|} \frac{w_i}{\sum_i w_i} 
	\log P(y_i \mid {\bf x}_i; \theta) 
	%  &= \min_\theta -\frac{1}{|\mathcal{D}|} \sum_{i=1}^{|\mathcal{D}|} 
	%    \frac{w_i}{\bar{w}} \log P(y_i \mid {\bf x}_i; \theta) 
	\label{eqn:emp_risk_resample} \\
	H(Y') 
	&\approx -\sum_{i=1}^{|\mathcal{D}|} \frac{w_i}{\sum_i w_i} 
	\log p'_{y_i}, 
	\label{eqn:entropy_resample}
	\end{align}
	where
	\begin{equation}
	p'_y = \frac{\sum_{i: y_i = y} w_i}{\sum_i w_i}.
	\end{equation}
	The goal is then to find the set of weights $\{w_i\}_{i=1}^{|\mathcal{D}|}$ that 
	minimizes the bias
	\begin{equation}
	\mathcal{B}(\mathcal{D}', \phi) = 1 - 
	\frac{\mathcal{R}^*(\mathcal{D}', \phi)}{H(Y')}.
	\end{equation}
	This leads to the optimization problem
	\begin{align}
	(w^*,\theta^*) &= \min_w \max_\theta \ \mathcal{V}(w, \theta) \label{eqn:minimax} \\
	\mathcal{V}(w, \theta) &= 1 - 
	\frac{\sum_i w_i \log P(y_i \mid {\bf x}_i; \theta)}
	{\sum_i w_i \log p'_{y_i}} \label{eqn:objective}
	\end{align}
	% This transforms \eqref{eqn:emp_risk} and \eqref{eqn:entropy} into
	% \begin{align}
	%   \mathcal{R}^*(\mathcal{D}', \phi) 
	%   &\approx \min_\theta - 
	%     \sum_{i=1}^{|\mathcal{D}|} p_i
	%     \log P(y_i \mid {\bf x}_i; \theta) 
	% %  &= \min_\theta -\frac{1}{|\mathcal{D}|} \sum_{i=1}^{|\mathcal{D}|} 
	% %    \frac{w_i}{\bar{w}} \log P(y_i \mid {\bf x}_i; \theta) 
	%     \label{eqn:emp_risk_resample} \\
	%   H(Y') 
	%   &\approx -\sum_{i=1}^{|\mathcal{D}|} p_i
	%     \log \sum_{i: y_i = y} p_i.
	%     \label{eqn:entropy_resample}
	% \end{align}
	% To guarantee that $p_i$ is a probability distribution over examples, 
	% we define it with a softmax function
	% \begin{equation}
	%   p_i = \rho_i({\bf w}) = \frac{e^{w_i}}{\sum_{i=1}^{|{\cal D}|} e^{w_i}}
	% \end{equation}
	% where ${\bf w} \in \mathbb{R}^{|{\cal D}|}$ is a vector of example dependant 
	% weights. The goal is then to find the set of weights 
	% $\{w_i\}_{i=1}^{|\mathcal{D}|}$ that minimizes the bias
	% \begin{equation}
	%   \mathcal{B}(\mathcal{D}', \phi) = 1 - 
	%   \frac{\mathcal{R}^*(\mathcal{D}', \phi)}{H(Y')}.
	% \end{equation}
	% This leads to the optimization problem
	% \begin{eqnarray}
	%   ({\bf w}^*,\theta^*) &=& \min_w \max_\theta  \mathcal{V}(w, \theta) \\
	%   \mathcal{V}({\bf w}, \theta) &=& 1 - 
	%     \frac{\sum_i \rho_i({\bf w}) \log P(y_i \mid {\bf x}_i; \theta)}
	%                              {\sum_i \rho_i({\bf w}) \log p'_{y_i}}
	%     \label{eqn:minimax}
	% \end{eqnarray}
	
	To solve the minimax game of \eqref{eqn:minimax}, we optimize the
	example weights  ${\bf w} = (w_1, \dots, w_{|\mathcal{D}|})$ and the bias
	estimator $\theta$ in an alternating fashion, similar to the procedure
	used to train adversarial networks \cite{goodfellow2014generative}.
	To guarantee that the weights $w_i$ are binary probabilities, we
	define $\bf w$ as the output of a sigmoid function
	$w_i = \rho(\omega_i) = (1 + e^{-\omega_i})^{-1} \in (0, 1)$, and
	update $\omega_i$ directly. Throughout the training iterations,
	the optimization of $\theta$ with a classification loss produces
	more accurate estimates of the representation bias. On the other
	hand, the optimization of $\bf w$ attempts to minimize this bias estimate
	by assigning larger weights to misclassified examples.
	Upon convergence, $\theta^*$ is a precise measure of the bias of the
	reweighted dataset, and $\bf w^*$ ensures that this bias is indeed minimized.
	
	Resampling ${\cal D}$ according to the distribution $w_i$ 
	leads to a dataset ${\cal D}'$  that is less biased for representation $\phi$, 
	while penalizing classes that do not contribute to the classification
	uncertainty. Because this has the effect of equalizing the preference of
	the dataset for different representations, we denote this
	process as dataset {\it REPresentAtion bIas Removal\/} (REPAIR).
	
	\begin{figure*}\RawFloats
		\centering
		\begin{minipage}[b]{.62\linewidth}
			\centering
			\begin{subfigure}[b]{\linewidth}
				\centering
				\includegraphics[width=\linewidth]{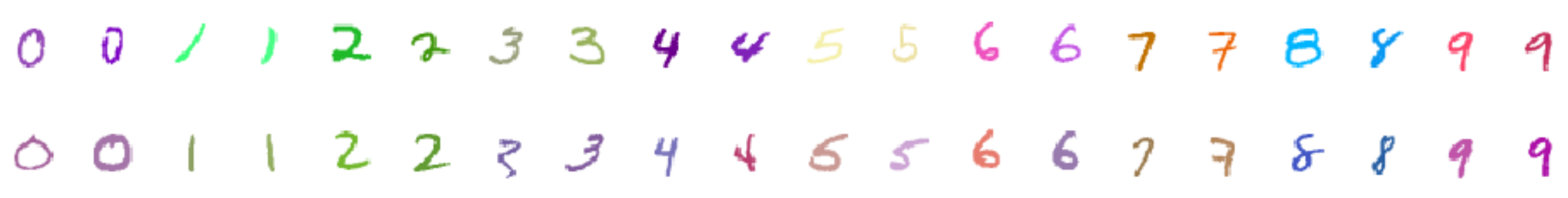}
				\caption{Random digit examples, before (\textbf{top}) and after
					(\textbf{bottom}) resampling.}
				\label{fig:colored_mnist}
			\end{subfigure} \\[1em]
			\begin{subfigure}[b]{0.38\linewidth}
				\centering
				\includegraphics[width=\linewidth]{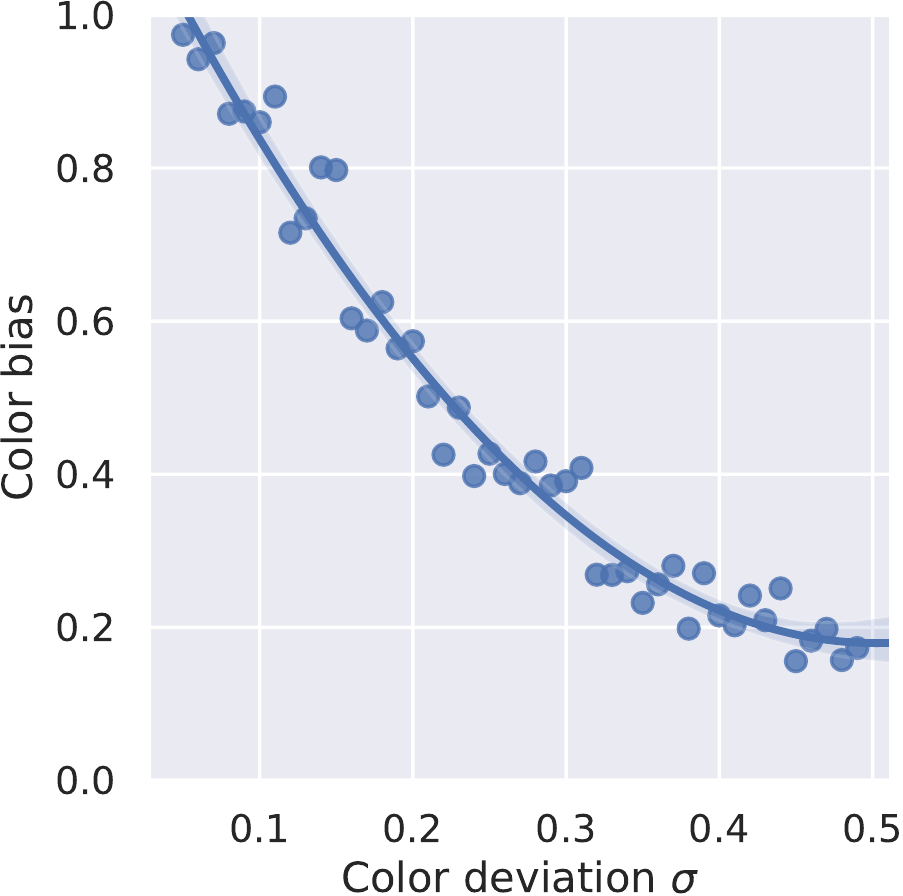}
				\caption{Controlling bias via intra-class color variation.}
				\label{fig:bias_std}
			\end{subfigure} \hspace{1em}
			\begin{subfigure}[b]{0.48\linewidth}
				\centering
				\includegraphics[width=\linewidth]{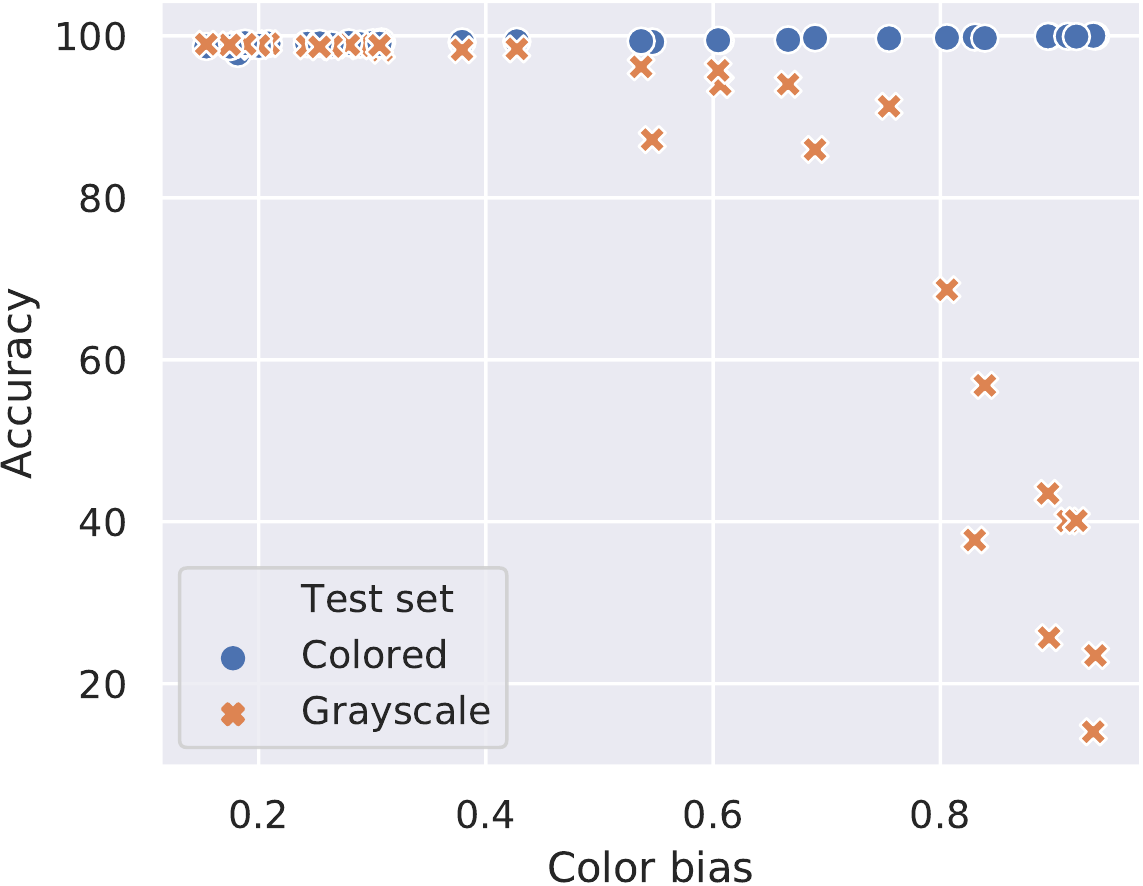}
				\caption{Test accuracy on biased (\textbf{colored}) and unbiased (\textbf{grayscale}) test sets.}
				\label{fig:colored_mnist_acc}
			\end{subfigure}
		\end{minipage} \hspace{1em}
		\begin{subfigure}[b]{0.3\linewidth}
			\centering
			\includegraphics[width=0.95\linewidth]{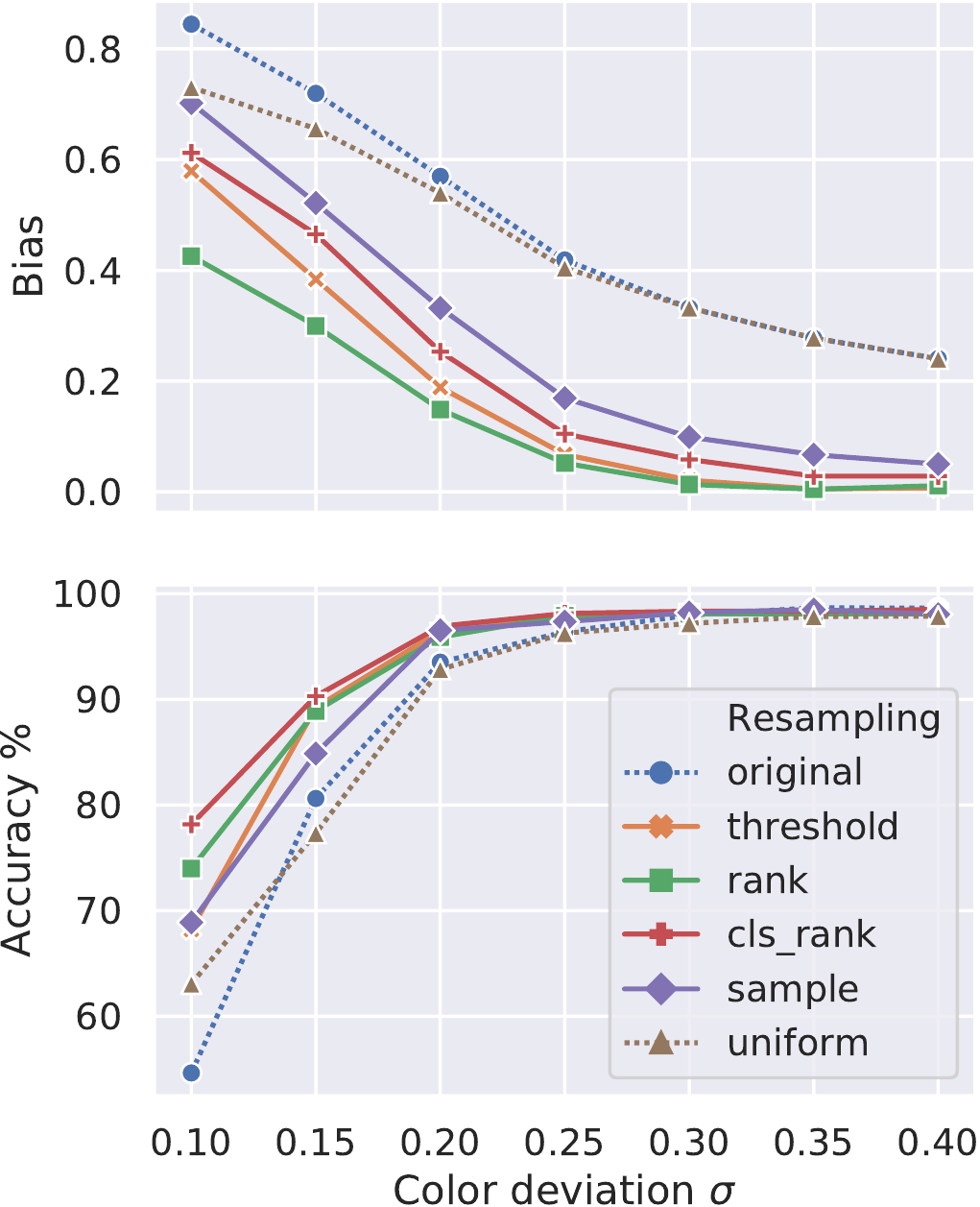}
			\caption{\textbf{Top:} Bias of resampled datasets. \textbf{Bottom:} Generalization performance.}
			\label{fig:color_bias_resampling}
		\end{subfigure} \vspace{.5em}
		\caption{Dataset Resampling on Colored MNIST Dataset.}
	\end{figure*}
	
	% \begin{figure*}\RawFloats
	%   \centering
	%   \begin{minipage}{0.8\linewidth}
	%     \includegraphics[width=\textwidth]{color_bias/colored_mnist_digits_cropped.pdf}
	%     \caption{Random digit examples, before (\textbf{top}) and after
	%       (\textbf{bottom}) resampling.}
	%     \label{fig:colored_mnist}
	%   \end{minipage}
	%   \begin{minipage}{.34\linewidth}
	%     \includegraphics[width=\linewidth]{color_bias/bias_std_scatter_cropped}
	%     \caption{Controlling bias via intra-class color variation.}
	%     \label{fig:bias_std}    
	%   \end{minipage}
	%   \begin{minipage}{.34\linewidth}
	%     \includegraphics[width=\linewidth,height=.8\linewidth]{color_bias/color_bias_acc_scatter_cropped}
	%     \caption{Test accuracy on biased (\textbf{colored}) and
	%       unbiased (\textbf{grayscale}) test sets.}
	%     \label{fig:colored_mnist_acc}
	%   \end{minipage}
	%   \begin{minipage}{.3\linewidth}
	%     \includegraphics[width=\linewidth,]{color_bias/bias_acc_resample_cropped}
	%     \caption{\textbf{Top:} bias of resampled datasets.
	%       \textbf{Bottom:} generalization performance.}
	%     \label{fig:color_bias_resampling}
	
	%   \end{minipage}
	%   %\caption{Dataset Resampling on Colored MNIST Dataset.}
	% \end{figure*}
	
	\subsection{Mini-batch Optimization}
	Efficient optimization on a large-scale dataset usually requires
	mini-batch approximations. The objective function above can be easily adapted
	to mini-batch algorithms. For this, it suffices to define
	\begin{equation}
	r_i = \frac{w_i}{\bar{w}} = \lvert\mathcal{D}\rvert \frac{w_i}{\sum_i w_i}
	\end{equation}
	where $\bar{w}$ is the sample average of $w_i$. The risk
	of \eqref{eqn:emp_risk_resample} and the entropy of
	\eqref{eqn:entropy_resample} can then be rewritten as
	\begin{align}
	\mathcal{R}^*(\mathcal{D}', \phi)
	&\approx \min_\theta -\frac{1}{|\mathcal{D}|} \sum_{i=1}^{|\mathcal{D}|} 
	r_i \log P(y_i \mid {\bf x}_i; \theta) \\
	H(Y') &\approx -\frac{1}{|\mathcal{D}|} \sum_{i=1}^{|\mathcal{D}|} r_i \log p'_{y_i}, 
	\end{align}
	and estimated from mini-batches, by replacing $|\mathcal{D}|$ with the mini-batch
	size. This enables the use of mini-batch SGD for solving the optimal
	weights of \eqref{eqn:minimax}. In practice REPAIR is performed on
	training and test splits of $\mathcal{D}$ combined, to ensure that the
	training and test sets distributions are matched after resampling.
	
	% While no term in the objective function explicitly encourages large weights, we argue that this formulation intrinsically ensures sufficient examples would be selected, as insufficient samples would lead to overfitting of the classifier and hence converge maximum bias. This is indeed verified in section \ref{sect:action}, where we visualize the weights learned on different datasets.
	
	\section{Case studies}
	
	In this section, we introduce two case studies for the study of bias
	reduction. The first is based on an artificial setting where bias can be
	controlled  explicitly. The second uses the natural setting of action 
	recognition from large-scale video datasets. While the ground-truth
	representation is not known for this setting, it is suspected that
	several biases are prevalent in existing datasets. In both
	cases, we investigate how representation biases can impair the fairness of 
	model evaluation, and prevent the learning of representations that 
	generalize well.
	
	\subsection{Colored MNIST} \label{sect:colored_mnist}
	
	The first case study is based on a modified version of 
	MNIST~\cite{lecun1998gradient}, which is denoted {\it Colored MNIST}.
	It exploits the intuition that digit recognition does not require
	color processing. Hence, the ground-truth representation for the {\it task\/}
	of digit recognition should not involve color processing. 
%	This is also guaranteed for the representation of highest performance,
%	commonly known as the {\it state of the art\/} representation,
	This is indeed guaranteed for representations learned
	on a grayscale dataset like MNIST.
	However, by introducing color, it is possible to create a dataset
	biased for color representations.
	
	\vspace{-1em}
	\paragraph{Experiment Setup.} To introduce \emph{color bias},
	we color each digit, using a different color for digits of
	different classes, as shown in Figure \ref{fig:colored_mnist}. 
	Coloring was performed by assigning to each example ${\bf x}_i$
	a color vector ${\bf z}_i = (r_i, g_i, b_i)$ in the RGB color space.
	Color vectors were sampled from class-dependent color distributions,
	\ie examples of digit $y$ were colored with vectors sampled from 
	a normal distribution distribution $p_y({\bf z})$ of mean 
	$\mu_y = (\mu^r_y, \mu^g_y, \mu^b_y)$ and covariance
	covariance $\Sigma_y = \sigma^2 I$. Since the simple observation of
	the color gives away the digit, Colored MNIST is biased for color
	representations $\bf z$. When learned on this dataset, a CNN can achieve
	high recognition accuracies without modeling any property of digits
	other than color. The color assignment scheme also enables
	control over the strength of this bias. By altering the means and 
	variances of the different classes, it is possible to create more
	or less overlap between the color distributions, making color more or less
	informative of the class label. 
	% The codes for implementing Colored MNIST environment and measuring the representation bias will be made publicly available for research works in the related field.
	% %  We experiment with different 
	% % values of $\sigma$ and observe how this affects CNN performance.
	% % {\color{red}What about Figure 1b). How is this obtained and what is
	% % it showing?}
	
	\vspace{-1em}
	\paragraph{Bias and Generalization.} To understand how representation bias 
	affects the fair evaluation of models, we trained a LeNet-5 CNN on the Colored 
	MNIST training set and compared its ability to recognize digits on the test
	sets of both the Colored MNIST and the original (grayscale) MNIST datasets. 
	To control the color bias of Colored MNIST, we varied the variance
	$\sigma$ of the color distributions. Figure~\ref{fig:bias_std} shows
	how the bias, computed with~\eqref{eqn:bias} on the colored test set, varies with $\sigma$.
	Clearly, increasing the variance $\sigma$ reduces bias. This was
	expected, since large variances create more overlap between the colors
	of the different classes, making color less discriminant.
	
	Figure \ref{fig:colored_mnist_acc} shows the recognition accuracy 
	of the learned CNN on the two test sets, as a function of the color bias. 
	A few observations can be drawn from the figure. 
	First, it is clear that CNN performance 
	on MNIST degrades as the bias increases. This shows that representation
	bias can hurt the generalization performance of the CNN. Second,
	this effect can be overwhelming. For the highest levels of bias, 
	the performance on MNIST drops close to chance level ($10\%$ on this 
	dataset). This shows that, when Colored MNIST is strongly biased for color,
	the CNN learns a representation that mostly accounts for color. While
	sensible to solve the training dataset (Colored MNIST), this is a terrible 
	strategy to solve the digit recognition {\it task\/} in general.
	As demonstrated by the poor performance on MNIST, the CNN has not learned
	anything about digits or digit recognition, simply overfitting to the bias of 
	the training set. Finally, and perhaps most important, this poor 
	generalization is not visible on the Colored MNIST test set, on which the 
	CNN reports deceptively high classification accuracy. The problem
	is that, like the training set, this is biased for color. Note 
	that adding more Colored MNIST style data will not solve the problem.
	The overfitting follows from the bias induced by the procedure used to
	{\it collect\/} the data, not from a {\it shortage\/} of data.
	Unless the dataset collection procedure is changed, adding more data
	only makes the CNN more likely to overfit to the bias.
	
	While this example is contrived, similar problems frequently
	occur in practice. A set of classes is defined and
	a data collection procedure, \eg data collection on the web, is chosen.
	These choices can introduce representation biases, which will be present
	independently of how large the dataset is. There are many possible
	sources of such biases, including the fact that some classes may 
	appear against certain types of backgrounds, contain certain objects,
	occur in certain types of scenes or contexts, exhibit some types of
	motion, \etc Any of these can play the role of  the digit colors
	of Colored MNIST. Since, in general, the test set is collected
	using a protocol similar to that used to collect the training set,
	it is impossible to detect representation bias from test set results
	or to reduce bias by collecting more data. Hence, there is a need
	for bias reduction techniques.
	
	% It is also worth noting that the unbiased test set is often inaccessible in real applications, where biases are implicit (rather than explicitly introduced in this experiment). This signifies the need for an unbiased evaluation procedure, such as building an unbiased test set, or diagnosing the sensitivity of models to different biases.
	
	\vspace{-1em}
	\paragraph{Resampling Strategies.} \label{sect:strategies}
	We next tested the ability of REPAIR to reduce representation bias
	on Colored MNIST. REPAIR was implemented according to \eqref{eqn:minimax} 
	on the colored training and test sets combined,
	with learning rates $\gamma_\theta = 10^{-3}$ and $\gamma_w = 10$ for 200 
	epochs, yielding an optimal weight vector ${\bf w}^*$ . 
	This was then used to implement a few sampling strategies.
	\begin{enumerate}
		\itemsep0em
		\item \textbf{Thresholding} (\texttt{threshold}): Keep all examples 
		$i$ such that $w_i \ge t$, where $t = 0.5$ is the threshold;
		\item \textbf{Ranking} (\texttt{rank}): Keep $p = 50\%$ examples 
		of largest weights $w_i$;
		\item \textbf{Per-class ranking} (\texttt{cls\_rank}): Keep 
		the $p = 50\%$ examples of largest weight $w_i$ from each class;
		\item \textbf{Sampling} (\texttt{sample}): Keep each example $i$ with probability $w_i$ (discard with probability $1 - w_i$).
		\item \textbf{Uniform} (\texttt{uniform}): Keep $p = 50\%$ examples uniformly at random.
	\end{enumerate}
	
	 \begin{figure*}\RawFloats
	     \centering
	     \includegraphics[height=0.25\linewidth]{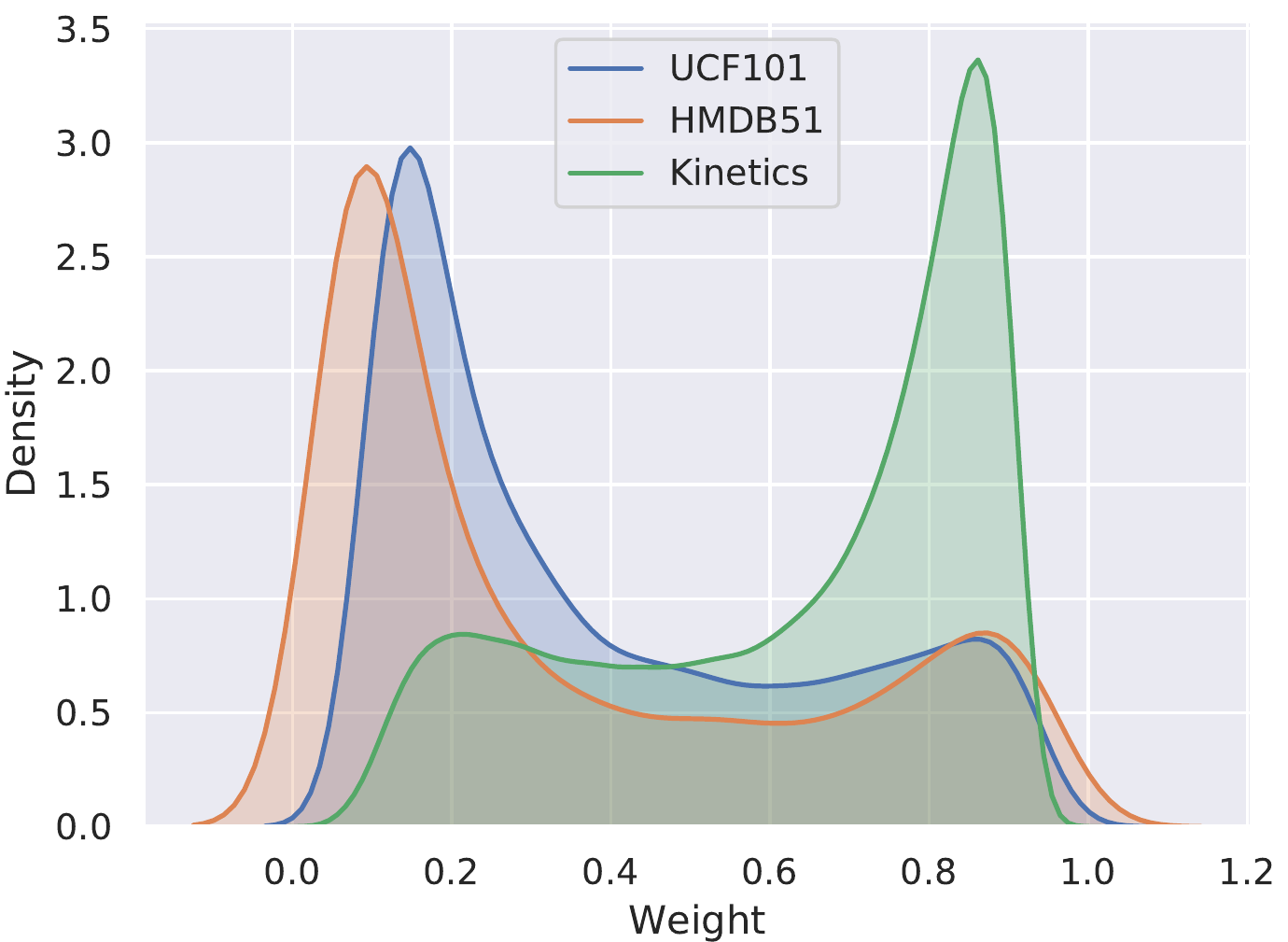} \hspace{2em}
	     \begin{subfigure}[b]{0.12\linewidth}
	     	\centering
	     	\includegraphics[width=\linewidth]{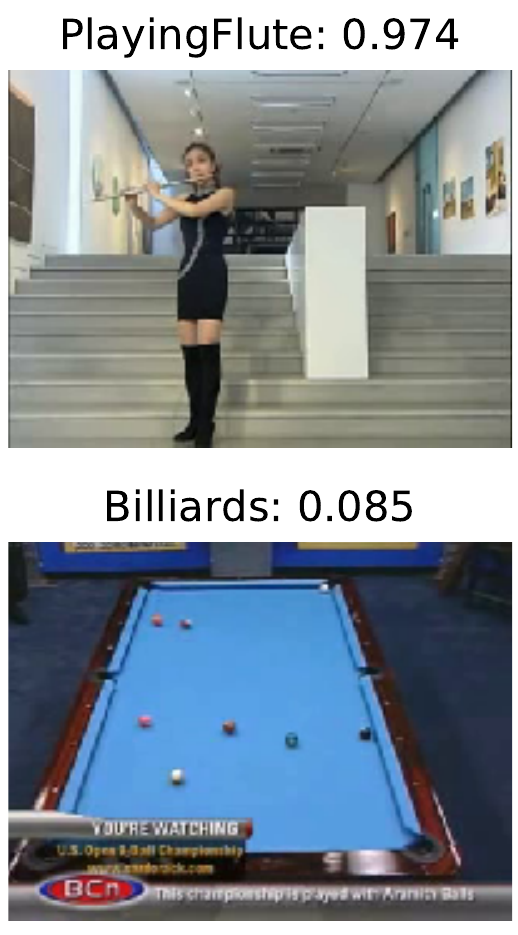}
	     	\caption{UCF101.}
	     \end{subfigure} \hspace{1em}
	     \begin{subfigure}[b]{0.125\linewidth}
		     \centering
		     \includegraphics[width=\linewidth]{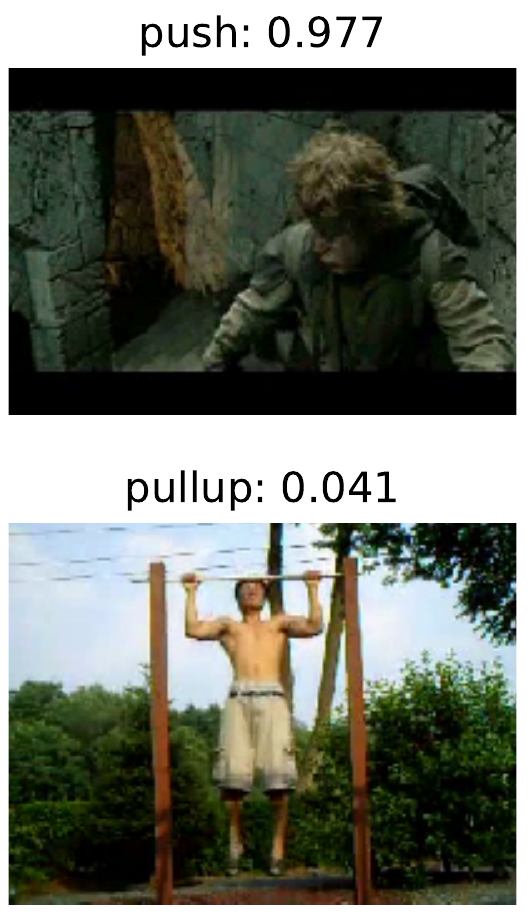}
		     \caption{HMDB51.}
		 \end{subfigure} \hspace{1em}
		 \begin{subfigure}[b]{0.125\linewidth}
			 \centering
			 \includegraphics[width=\linewidth]{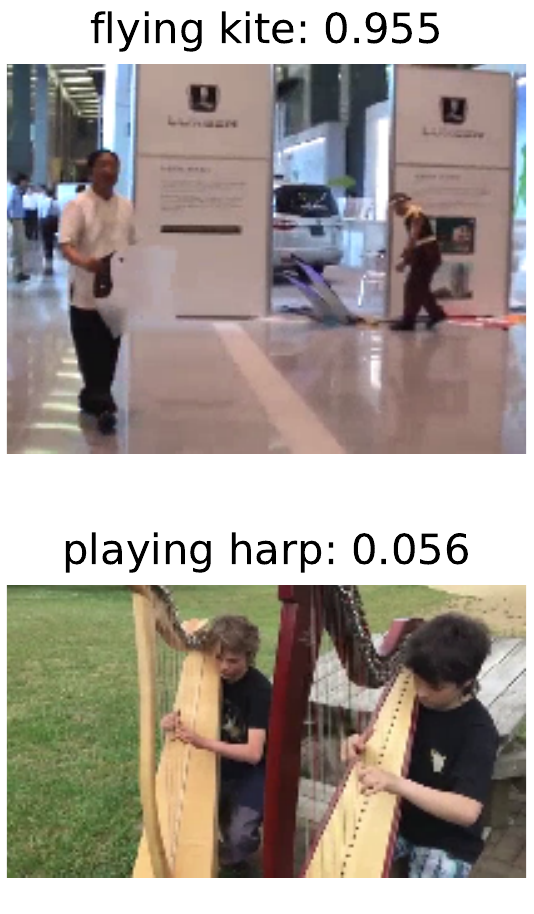}
			 \caption{Kinetics.}
		 \end{subfigure} \vspace{.5em}
	 \caption{\textbf{Left}: Histograms of resampling weights. \textbf{Right}: Examples with highest and lowest weights from each dataset.}
	 \label{fig:resample_hist}
	\end{figure*}
	
%	\begin{figure*}[t]\RawFloats
%		\centering
%		\begin{tabular}{cccc}
%			\multirow{2}{*}{\includegraphics[height=0.27\textwidth]{static_bias/resample_hist_cropped}}& {\bf UCF101} & {\bf HMDB51} & {\bf Kinetics} \\
%			& \includegraphics[height=0.23\textwidth]{static_bias/ucf101_frames_cropped}
%			& \includegraphics[height=0.23\textwidth]{static_bias/hmdb51_frames_cropped}
%			& \includegraphics[height=0.23\textwidth]{static_bias/kinetics_frames_cropped}
%		\end{tabular}
%		\caption{\textbf{Left}: Histograms of resampling weights.
%			\textbf{Right}: Examples with highest (top)  and lowest (bottom)
%			weights on each dataset.}
%		\label{fig:resample_hist}
%	\end{figure*}
	
	To evaluate the resampling strategies, we tested their ability to reduce 
	representation bias and improve model generalization 
	(test accuracy on MNIST). The experiments were performed with different
	color variances $\sigma$, to simulate different level of bias.
	The results were averaged over 5 runs under each setting. 
	% 
	% VERSION 1: BIAS REDUCTION & GENERALIZATION IMPROVEMENT
	% Bias reduction is measured by the difference between bias of the Colored MNIST dataset $\cal D$ and its REPAIRed version $\cal D'$. Figure \ref{fig:color_bias_resampling} (\textbf{top}) shows this quantity as a function of $\sigma$. All four strategies guided by resampling weights $w_i$ led to a significant reduction in color bias, relative to both the bias before resampling and that achieved by uniform resampling. Among them, thresholding and ranking proved more effective for large biases (small values of $\sigma$).
	% {\color{red} This makes no sense. How can the bias decrease? Why doesn't the
	% CNN just choose to do grayscale?} 
	% 
	% VERSION 2: BIAS & GENERALIZATION
	Figure \ref{fig:color_bias_resampling} (\textbf{top}) shows the bias after
	resampling, as a function of $\sigma$. All four strategies where resampling
	leverages the weights $w_i$ led to a significant reduction in color bias,
	relative to both the bias before resampling 
	% {\color{red} can we show the bias before resampling as well?}
	and that achieved by uniform
	resampling. Among them, thresholding and ranking were more effective for large biases (small values of $\sigma$). 
	The reduction in color bias also led to better model generalization, as
	shown in Figure \ref{fig:color_bias_resampling} (\textbf{bottom}).
	This confirms the expectation that large bias harms the generalization
	ability of the learned model. Visual inspection of examples from the
	REPAIRed dataset, shown in Figure \ref{fig:colored_mnist} (\textbf{bottom}),
	explains this behavior. Since it becomes harder to infer the digits from their
	color, the CNN  must rely more strongly on shape modelling, and thus
	generalizes better.

	% \begin{figure}
	%     \centering
	%     \includegraphics[width=0.45\textwidth, trim={0 5mm 0 15mm}, clip]{color_bias/bias_acc_resample}
	%     % \includegraphics[width=0.45\textwidth, trim={0 5mm 0 15mm}, clip]{color_bias/bias_acc_diff_resample}
	% \caption{Resampling strategies on Colored MNIST. *Not finalized}
	% \caption{Bias reduction of resampled datasets (\textbf{top}), and gain in generalization performance they provide (\textbf{bottom}).}
	% \caption{Bias of resampled datasets (\textbf{top}), and generalization performance they provide (\textbf{bottom}).}
	% \label{fig:color_bias_resampling}
	%\end{figure}
	
	\subsection{Scenario II: Action Recognition} \label{sect:action}
	Video action recognition is a complex task with various potential sources
	of bias, as shown by the analysis of \cite{li2018resound}. In this work,
	we focus on static bias, \ie bias towards single-frame representations.
	The observation that popular action recognition datasets like
	UCF101 \cite{soomro2012ucf101} and Kinetics \cite{kay2017kinetics} are biased
	for static features, in that substantial portions of their
	data can be solved without leveraging temporal information, has
	been reported by several recent works \cite{huang2018makes,
		feichtenhofer2018have}. Yet, little attention has been given to
	the impact of bias on learning and evaluation of
	action recognition models.
	
	In this section, we present an in-depth analysis on the connection
	between static dataset bias and model performance on the dataset.
	We used REPAIR to manipulate the static bias of a dataset, through
	the selection of examples according to their learned weights. We then evaluated
	how the performance of prevailing action recognition models changes as a
	function of static bias. This allowed us to compare the sensitivity of
	the models to the presence of  static cues in the data.
	Finally, by examining models trained on datasets with different level of
	static bias, we assessed their ability to capture temporal information and
	learn human actions that generalize across datasets.
% 	All REPAIRed datasets used in this work \textcolor{red}{will be publicly available\footnote{\texttt{https://github.com/JerryYLi/Dataset-REPAIR/}},
% 	to enable comparisons of new algorithms at different levels of static
% 	dataset bias}.
	
	\begin{figure}\RawFloats
		\centering
		\includegraphics[width=.78\textwidth]{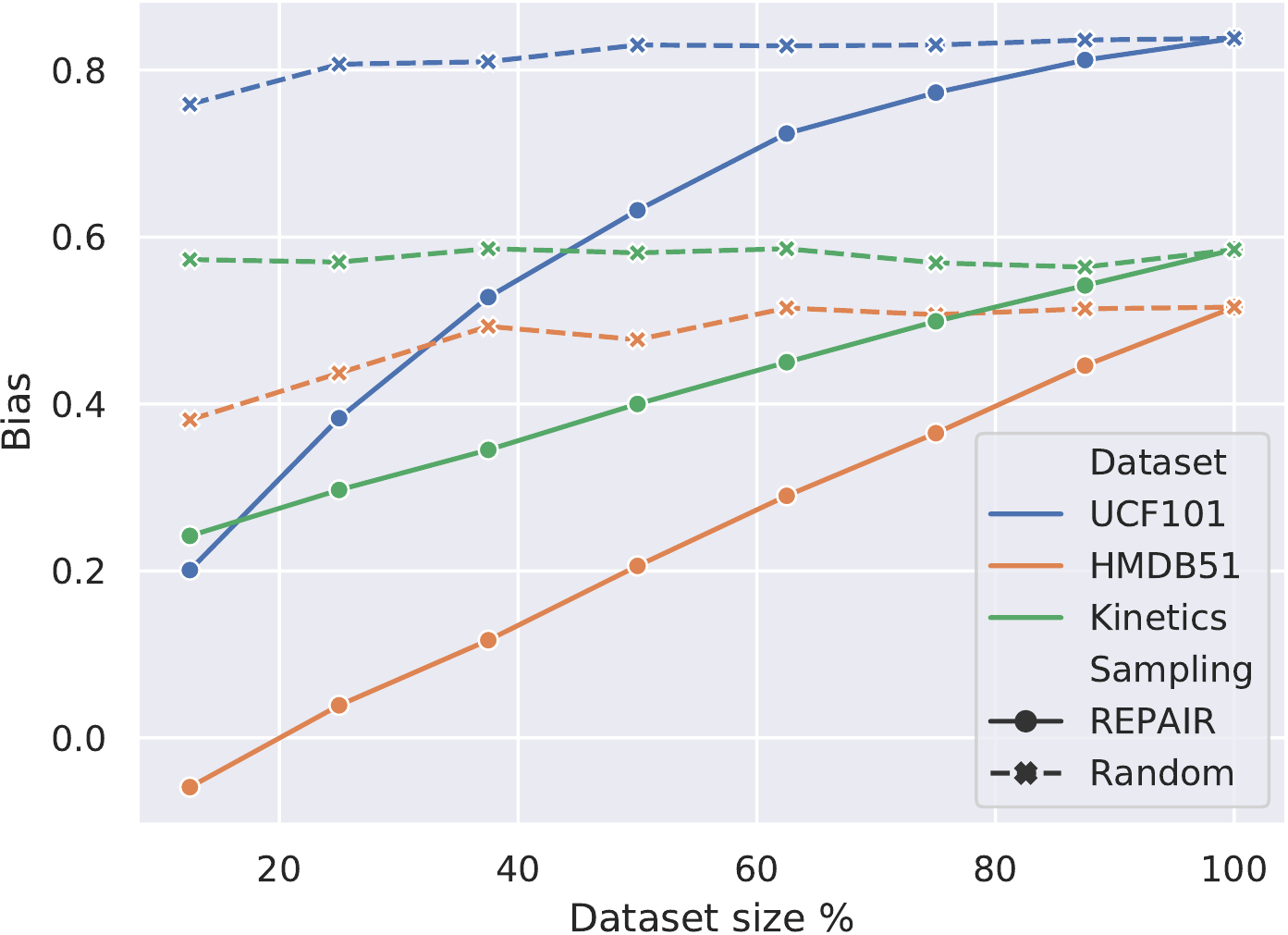}
		\vspace{.5em}
		\caption{Static bias as a function of dataset size.
			Examples are removed either randomly or according to their weights.}
		\label{fig:bias_dset_size}
	\end{figure}

	\begin{figure*}\RawFloats
	\centering
	\includegraphics[height=0.255\textwidth]{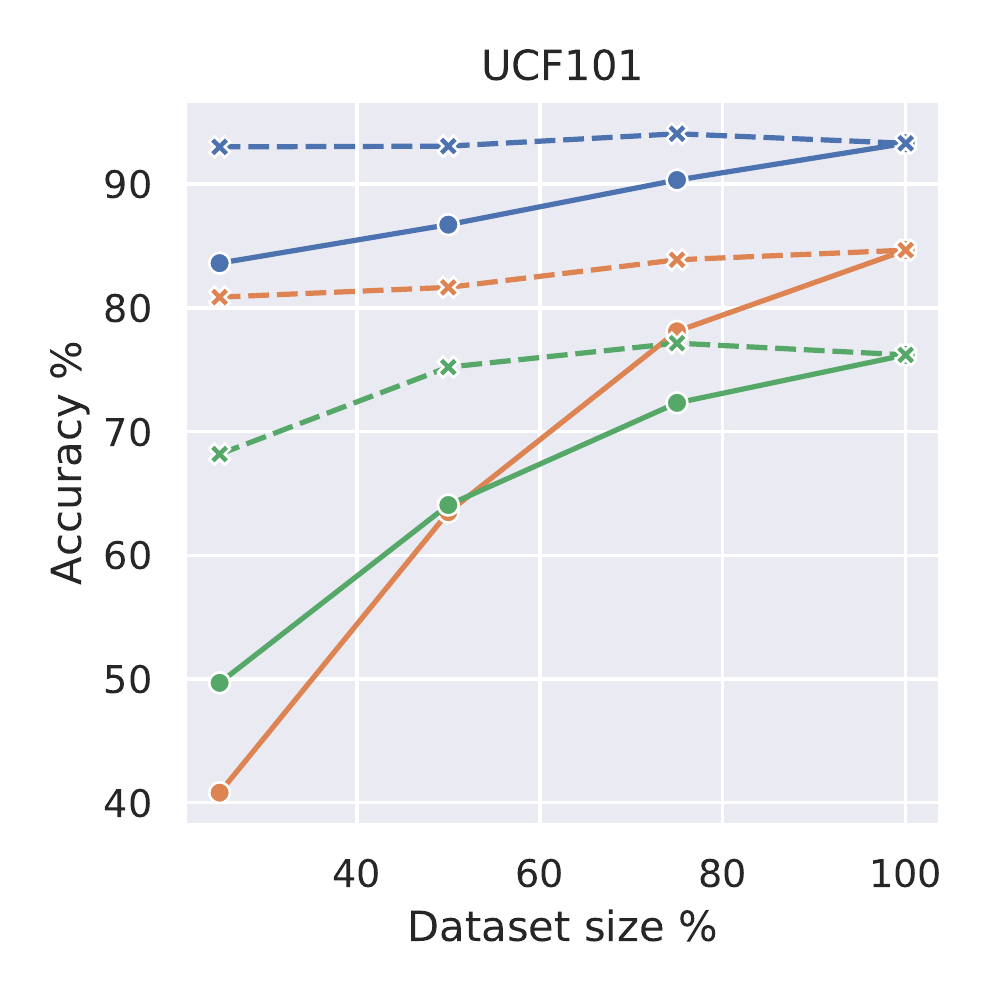}
	\includegraphics[height=0.255\textwidth]{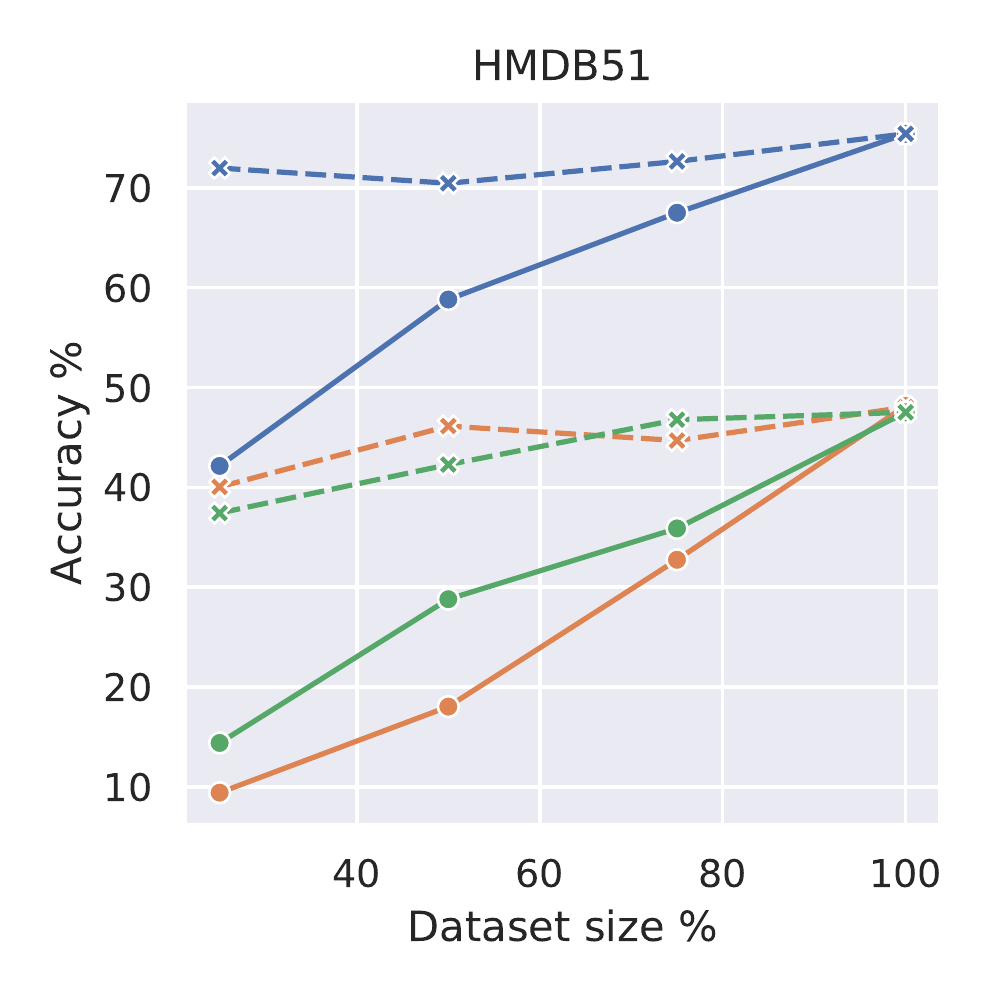}
	\includegraphics[height=0.255\textwidth]{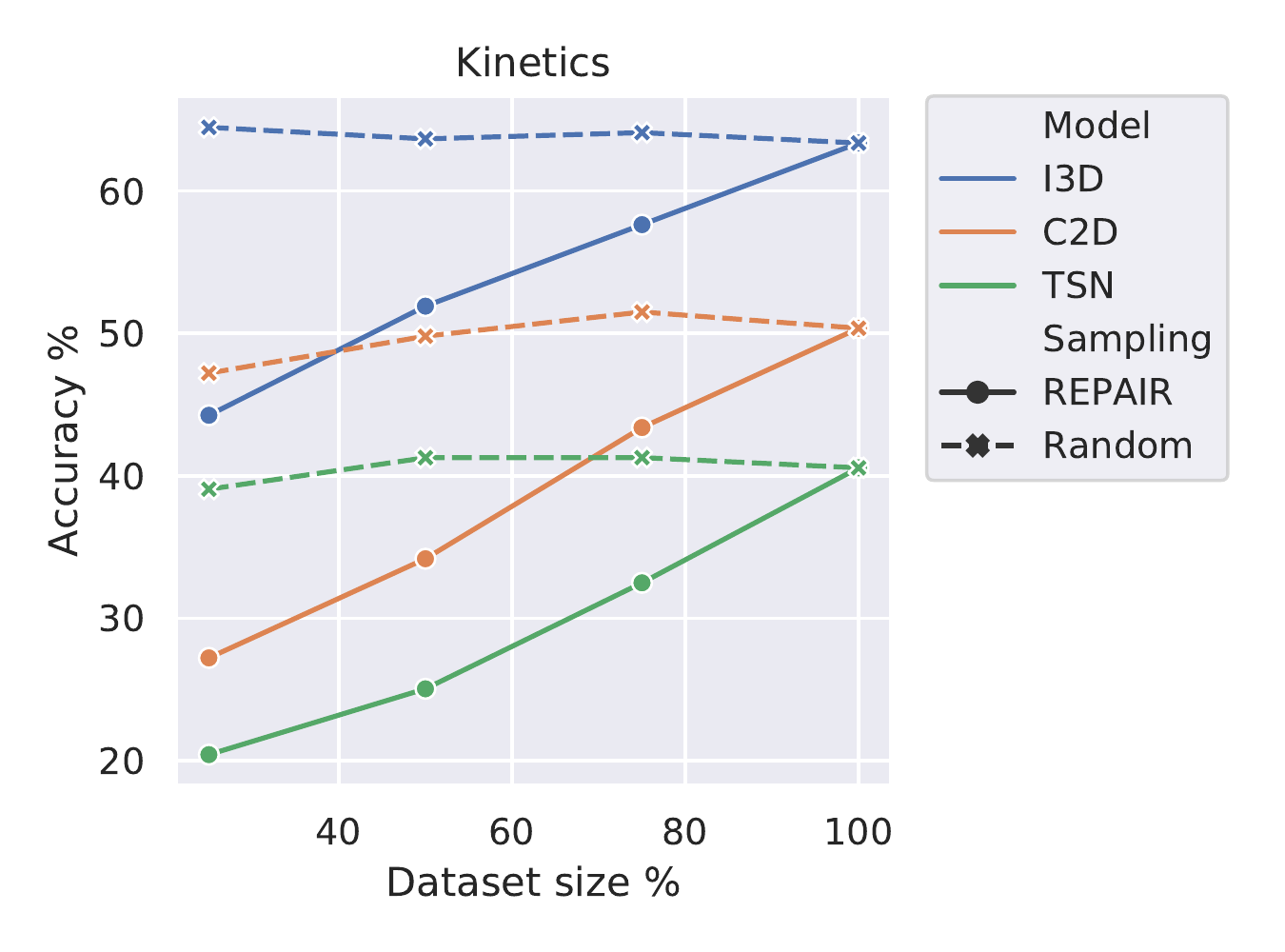}
	\caption{Evaluations of action recognition models on resampled datasets.}
	\label{fig:action_bias_accs}
	\end{figure*}
	
	\vspace{-1em}
	\paragraph{Static Bias Minimization.}
	We implemented $\phi$ with ImageNet features extracted from the
	ResNet-50 \cite{he2016deep}, a typical representation for static
	image recognition. REPAIR weights were learned for 20k iterations with
	learning rate $\gamma_\theta = 10^{-3}$ and $\gamma_w = 10^{-3} |\mathcal{D}|$,
	as the number of weights $w_i$ to be learned grows linearly with dataset size. 
	Figure \ref{fig:resample_hist} (\textbf{left}) shows the distribution of
	resampled weights learned for UCF101 \cite{soomro2012ucf101},
	HMDB51 \cite{kuehne2011hmdb} and Kinetics \cite{kay2017kinetics};
	A random frame from videos of highest and lowest weights is displayed in
	Figure \ref{fig:resample_hist} (\textbf{right}).
	Several observations can be made. First, REPAIR uncovers videos with
	abundant static cues (\eg pool tables in \textit{billiards} and parallel
	vertical lines in \textit{playing harp}). These videos receive lower scores
	during resampling. On the other hand, videos with no significant static
	cues (\eg complex human interactions in \textit{push}), are more likely
	to be selected into the resampled dataset. Second, the optimization does not
	learn the trivial solution of setting all weights to zero. Instead, the
	weights of all videos range widely from 0 to 1, forming two clusters at
	both ends of the histogram. Third,  while all datasets contain a substantial
	amount of videos that contribute to static bias, Kinetics contained
	more videos of large weight ($w > 0.5$), enabling more freedom in
	the assembly of new datasets.
	
	Following the \textit{ranking} strategy of section \ref{sect:strategies},
	the videos were sorted by decreasing weights. Resampled datasets were
	then formed by keeping the top $p\%$ of the data
	and eliminating the rest (value of $p$ varies). Figure \ref{fig:bias_dset_size} shows how the
	static biases of the three datasets are reduced by this resampling
	procedure. This is compared to random sampling the same number of examples.
	The bias of \eqref{eqn:bias} was computed as the maximum over 5 measurements,
	each time training the bias estimator $\theta$ with a different weight
	decay, ranging from $10^{-1}$ to $10^{-5}$, so as to prevent overfitting due
	to insufficient training data. The bias curves validate the effectiveness
	of REPAIR, as the static classifier performs much weaker
	on the REPAIRed datasets (hence less static bias). This
	is unlike random sampling, which does not affect the bias measurements
	significantly. These results are also interesting because they enable
	us to alter static dataset bias within a considerable range of values,
	for further experiments with action recognition models.
	
	\vspace{-1em}
	\paragraph{Video Models \vs Static Bias.}
	To evaluate how representation bias affects the action recognition
	performance of different models, we trained and evaluated three models
	from the literature on the original and REPAIRed action datasets:
	\begin{enumerate}
		\itemsep0em
		\item 2D ConvNet (\textbf{C2D}): Baseline ResNet-50 applied
		independently to each frame, predictions then averaged. Pre-trained on
		ImageNet \cite{russakovsky2015imagenet}.
		\item Temporal segment network (\textbf{TSN}) \cite{wang2016temporal}:
		Aggregating features (we used \textit{RGB-diff}) from multiple snippets of the video according to their
		segmental consensus. Pre-trained on ImageNet.
		\item Inflated 3D ConvNet (\textbf{I3D}) \cite{carreira2017quo}:
		Spatiotemporal convolutions inflated from a 2D Inception-v1 network.
		Pre-trained on ImageNet and Kinetics.
	\end{enumerate}
	
	The networks were fine-tuned through SGD with 
	learning rate $10^{-3}$ and momentum 0.9, for 10k iterations on UCF101 
	and HMDB51 and 50k iterations on Kinetics. Figure \ref{fig:action_bias_accs}
	shows the performance of all three models on the three datasets.
	It is clear that all networks have weaker performance on the
	REPAIRed datasets (smaller static bias) than on the origonal ones.
	%This is understandable as the models might have benefited from their 
	%ImageNet pre-training in classifying the videos.
	The drop in accuracy is a measure of the reliance of the action models on
	static features, which we denote as the static bias dependency of
	the models. More precisely, we define the {\it static bias dependency
		coefficient\/} $\beta$ of a model on representation $\phi$ as the difference
	between model performance on randomly sampled and REPAIRed datasets,
	averaged over resampling rates (0.25, 0.5 and 0.75 in this case).
	% the ratio between its performance on randomly sampled and REPAIRed datasets, averaged over different resampling rates (equation?). 
	The larger $\beta$ is, the more the model leverages static bias to
	solve the dataset; $\beta = 0$ indicates that model performance is
	independent of static bias. Table \ref{tab:model_bias} summarizes
	the dependency coefficients of the different models, showing
	that C2D has much larger static bias dependency than TSN and I3D.
	While this comparison is not, by itself, enough to conclude that one model
	is superior to the rest, the reduced static bias dependency of the more
	recent networks suggests that efforts towards building better spatiotemporal
	models are paying off.
	% which we consider to be the \emph{static bias of models}. It is therefore 
	% natural to quantify the bias of a model on dataset $\mathcal{D}$ by its 
	% average performance drop on the resampled versions of $\mathcal{D}$ versus 
	% random sampling. We show the results on Table \ref{tab:model_bias}, clearly 
	% depicting the high static bias of C2D relative to TSN and I3D. While we 
	% shall not conclude merely from this comparison that one model is superior 
	% to the rest, the decreasing bias of models suggests that recent efforts on 
	% building spatiotemporal networks have indeed paid off.

	Another notable observation from Figure
	\ref{fig:action_bias_accs} is that the ranking of models by their performance
	on the original dataset is not necessarily meaningful.
	For example, while C2D outperforms TSN on UCF101, the reverse holds
	after 50\% and 25\% resampling. This shows that rankings of action
	recognition architectures could simply reflect how much
	they leverage representations biases.
	For example, stronger temporal models could underperform
	weaker static models if the dataset has a large static bias, potentially
	leading to unfairness in model evaluation. By reducing
	representation bias, REPAIR can alleviate this unfairness.
	
	\begin{table}\RawFloats
		\centering
		%   \footnotesize
		\scalebox{0.95}{
			% \begin{tabular}{c|ccc}
			%      & C2D \cite{simonyan2014two} & TSN \cite{wang2016temporal} & I3D \cite{carreira2017quo} \\ \hline \hline
			%     UCF101 & 21.34\% & 11.49\% & \textbf{6.50\%} \\
			%     HMDB51 & 23.56\% & \textbf{14.84\%} & 15.54\% \\
			%     Kinetics & 14.57\% & 14.55\% & \textbf{12.80\%} \\ \hline
			%     Average & 19.82\% & 13.63\% & \textbf{11.61\%}
			% \end{tabular}
			\begin{tabular}{c|ccc}
				& C2D \cite{simonyan2014two} & TSN \cite{wang2016temporal} & I3D \cite{carreira2017quo} \\ \hline \hline
				UCF101 & 0.213 & 0.115 & \textbf{0.065} \\
				HMDB51 & 0.236 & \textbf{0.148} & 0.155 \\
				Kinetics & 0.146 & 0.146 & \textbf{0.128} \\ \hline
				Average & 0.198 & 0.136 & \textbf{0.116}
			\end{tabular}
		}
		\vspace{.5em}
		\caption{Static bias dependency coefficient $\beta$ of the three action recognition
			models, evaluated on the three different datasets.}
		\label{tab:model_bias}
	\end{table}
	
	% {\color{red} Is there another concrete result that comes out of all this?}
	
	\begin{figure*}\RawFloats
		\centering
		\includegraphics[width=0.95\textwidth]{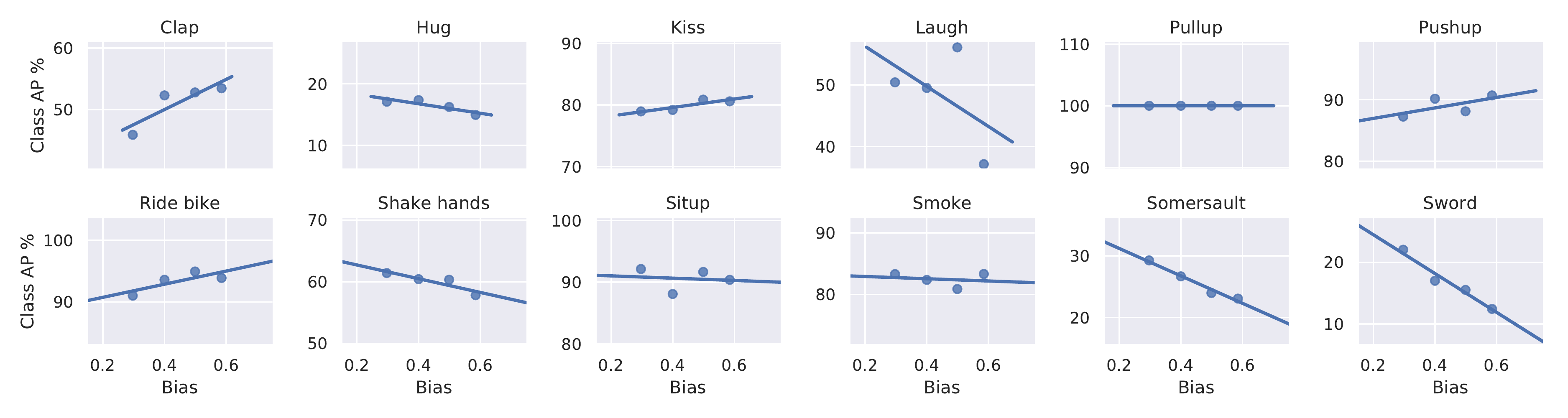}
		\caption{Class-level cross-dataset generalization of I3D models
			trained on REPAIRed \textbf{Kinetics} datasets. Test set is
			\textbf{HMDB51}.}
		\label{fig:generalization}
	\end{figure*}
	
	\begin{table}[t]\RawFloats
        \centering
        \scalebox{0.95}{
        \begin{tabular}{c|cccc}
            Dataset size & 100\% (\emph{orig.}) & 75\% & 50\% & 25\% \\ \hline
            mAP \% & 61.48 & \textbf{63.45} & 63.06 & 63.24
        \end{tabular}} \vspace{1mm}
        \caption{Cross-dataset generalization from \textbf{Kinetics} to \textbf{HMDB51} over 12 common classes. See Figure \ref{fig:generalization} for per-class AP.}
        \label{tab:generalization}
    \end{table}
	
	\vspace{-1em}
	\paragraph{Cross-dataset Generalization.}
	We next compared the performance of the I3D models trained on the
	original and resampled datasets. Unlike the Colored MNIST experiment of
	Figure~\ref{fig:colored_mnist_acc}, it is not possible to evaluate
	generalization on an unbiased test set. Instead, we measured
	cross-dataset generalization, with similar setup to \cite{torralba2011unbiased}. 
	This assumes that the datasets do
	not have the exact same type of representation bias, in which case
	overfitting to the biases of the training set would hamper generalization
	ability.
	
	% We follow a similar procedure to \cite{torralba2011unbiased} in the experiment setup, except that we focus on 12 common classes between HMDB51 and Kinetics, and train multi-class instead of binary classifiers. Despite that more common action classes can be found between UCF101 and Kinetics, we note that both datasets are collected from YouTube videos and could share very similar distributions, negating the original intention of testing cross-dataset generalization. HMDB51, on the contrary, consists of videos sourced from movies and other public databases, making it more suitable for this experiment.
% 	
% 	We adopted a setup similar to that of \cite{torralba2011unbiased}.
% 	We started by selecting $12$ classes shared by HDMB51 and Kinetics.
	We used Kinetics as the training set and
	HMDB51 as the test set for generalization performance. 
	The two datasets had 12 action classes in common.
	While more classes are shared among UCF101 and Kinetics, they are
	both collected on YouTube and have very similar distributions.
	HMDB51, on the contrary, consists of videos sourced from movies and other
	public databases and poses a stronger generalization challenge. 
	The I3D models were trained on the 12 classes of the original and REPAIRed
	versions of Kinetics, and evaluated \textit{without fine-tuning} on
	the same classes of HMDB51. Model generalization was evaluated by
	average precision (AP), measured for each of the
	common classes.
	
	Figure \ref{fig:generalization} summarizes the generalization performance
	of the I3D models as a function of the static bias in the
	\textit{training} set, for each of the $12$ classes.
	To visualize the correlation among the two
	variables we also show a line regressed on the different points.
	The four points in each subplot, from \textit{right} to \textit{left}, correspond to models trained on the original dataset and the REPAIRed ones with 75\%, 50\% and 25\% sampling rate, respectively.
	Of the 12 classes, 7 showed a negative correlation between bias and
	generalization. Furthermore, the correlation tends to
	be strongly negative for the classes where the model generalizes the worst,
	namely \textit{hug}, \textit{somersault} and \textit{sword}. On the
	contrary, positive correlation occurs on the classes of high
	generalization performance. This indicates that, at the class level,
	there are strong differences between the biases of the two datasets.
	Classes that generalize well are those where biases are shared across
	datasets, while low performance ones have different biases.
	% The classes that benefit from higher bias values, like \textit{pushup} and \textit{ride bike}, contain more static cues than others.
	The mean average precision (\textit{mAP}) of all 12 classes increased by $\sim $2\% after resampling as shown in Table \ref{tab:generalization}, validating the effectiveness of REPAIR on improving model generalization. 
%	This experiment shows that reducing bias
%	affects mostly the performance on classes where biases are different across datasets, which
%	are the ones that hurt generalization. Hence, dataset REPAIR
%	enhances the generalization ability of the algorithms learned on a
%	dataset.
	
	\vspace{-1em}
	\paragraph{Temporal Reasoning in Learned Models.}
	Finally, we analyzed in greater detail the I3D models learned on the REPAIRed
	datasets, aiming to understand the improvement in their generalization
	performance. We hypothesize that, with less static cues to hold on to,
	the network (even with unchanged structure) should learn to make inferences
	that are more dependent on the temporal structure of the video.
	To test this hypothesis, we performed a simple experiment. Given an
	input video, we measured the Euclidean distance between the feature vectors
	extracted from its regular 64-frame clip and its time reversed version.
	This distance was averaged over all video clips in the test set, and
	is denoted as the {\it temporal
		structure score\/} of the model.
	Larger scores reflect the fact that the model places more emphasis on the
	temporal structure of the video, instead of processing frames individually.
	% measuring the model's distinction of pure temporal information.
	Note that, because the 3D convolution kernels of I3D are initialized by
	duplicating the filters of a 2D network \cite{carreira2017quo}, the
	temporal structure score should be zero in the absence of training. 
	
	For this experiment, we used the test set of the 20BN-Something-Something-V2
	\cite{mahdisoltani2018fine} dataset, which is known for the fact that its
	action classes are often dependent on the arrow of time (\eg \textit{opening} \vs
	\textit{closing}, or \textit{covering} \vs \textit{uncovering}).
	Table \ref{tab:time_reversal} summarizes the scores obtained for all learned
	models on the test set of Something-Something.
	The table shows that, for REPAIRed datasets, the score increases as more
	biased videos are removed from the dataset. This is not a mere consequence
	of reduced dataset size, since the score varies
	little for random discarding of the same number of examples.
	This is evidence that static bias is an obstacle to the modeling of video
	dynamics, and dataset REPAIR has the potential to overcome this obstacle.
	
	\begin{table}\RawFloats
		\centering
%		\footnotesize
		% \begin{tabular}{cc|cccc}
		%     \multirow{2}{*}{Training set} & \multirow{2}{*}{Sampling} & \multicolumn{4}{c}{Training set size} \\
		%     & & 100\% & 75\% & 50\% & 25\% \\ \hline \hline
		%     \multirow{2}{*}{UCF101} & REPAIR & \multirow{2}{*}{1.75} & 1.76 & 1.92 & \textbf{1.96} \\
		%      & Random & & 1.78 & 1.82 & 1.78 \\ \hline
		%     \multirow{2}{*}{HMDB51} & REPAIR & \multirow{2}{*}{2.04} & 2.03 & 2.25 & \textbf{2.31} \\
		%      & Random & & 2.01 & 2.17 & 2.09 \\ \hline
		%     \multirow{2}{*}{Kinetics} & REPAIR & \multirow{2}{*}{3.65} & 3.63 & 3.68 & \textbf{3.83} \\
		%      & Random & & 3.66 & 3.56 & 3.57 \\
		% \end{tabular}
		\scalebox{0.75}{
		\begin{tabular}{cc|cccc}
			\multirow{2}{*}{Training set} & \multirow{2}{*}{Sampling} & \multicolumn{4}{c}{Training set size} \\
			& & 100\% & 75\% & 50\% & 25\% \\ \hline \hline
			\multirow{2}{*}{UCF101} & REPAIR & \multirow{2}{*}{1.76} & \textbf{1.76} & \textbf{1.92} & \textbf{1.96} \\
			& Random & & 1.75 $\pm$ .03 & 1.79 $\pm$ .04 & 1.78 $\pm$ .05 \\ \hline
			\multirow{2}{*}{HMDB51} & REPAIR & \multirow{2}{*}{2.04} & \textbf{2.03} & \textbf{2.25} & \textbf{2.31} \\
			& Random & & 2.02 $\pm$ .02 & 2.07 $\pm$ .07 & 2.08 $\pm$ .02 \\ \hline
			\multirow{2}{*}{Kinetics} & REPAIR & \multirow{2}{*}{3.67} & 3.63 & \textbf{3.68} & \textbf{3.83} \\
			& Random & & \textbf{3.66} $\pm$ .08 & 3.56 $\pm$ .04 & 3.59 $\pm$ .03 \\
		\end{tabular}
		}
		\vspace{.5em}
		\caption{Temporal structure scores of I3D models trained on UCF101, HMDB51, and Kinetics, evaluated on the Something-Something-V2 test set.}
		\label{tab:time_reversal}
	\end{table}
	
	\section{Conclusion}
		We presented \textit{REPresentAtion bIas Removal} (REPAIR), a novel dataset resampling procedure for minimizing the representation bias of datasets. Based on our new formulation of bias, the minimum-bias resampling was equated to a minimax problem and solved through stochastic gradient descent.
		Dataset REPAIR was shown to be effective, both under controlled settings of Colored MNIST and in large-scale modern action recognition datasets. 
		We further introduced a set of experiments for evaluating the effect of bias removal, which relates representation bias to the generalization capability of recognition models and the fairness of their evaluation.
		We hope our work will motivate more efforts on understanding and addressing the representation biases in different areas of machine learning.
	
	\vspace{-1em}
	\paragraph{Acknowledgement}
	    This work was partially funded by NSF awards IIS-1546305 and IIS-1637941, and NVIDIA GPU donations.

	{\small
		\bibliographystyle{ieee_fullname}
		\bibliography{references}

\begin{thebibliography}{10}\itemsep=-1pt

\bibitem{anne2018women}
Lisa Anne~Hendricks, Kaylee Burns, Kate Saenko, Trevor Darrell, and Anna
  Rohrbach.
\newblock Women also snowboard: Overcoming bias in captioning models.
\newblock In {\em European Conference on Computer Vision (ECCV)}, pages
  771--787, 2018.

\bibitem{beery2018recognition}
Sara Beery, Grant Van~Horn, and Pietro Perona.
\newblock Recognition in terra incognita.
\newblock In {\em European Conference on Computer Vision (ECCV)}, pages
  472--489, 2018.

\bibitem{bolukbasi2016man}
Tolga Bolukbasi, Kai-Wei Chang, James~Y Zou, Venkatesh Saligrama, and Adam~T
  Kalai.
\newblock Man is to computer programmer as woman is to homemaker? debiasing
  word embeddings.
\newblock In {\em Advances in Neural Information Processing Systems (NIPS)},
  pages 4349--4357, 2016.

\bibitem{carreira2017quo}
Joao Carreira and Andrew Zisserman.
\newblock Quo vadis, action recognition? a new model and the kinetics dataset.
\newblock In {\em Conference on Computer Vision and Pattern Recognition
  (CVPR)}, pages 4724--4733, 2017.

\bibitem{chawla2002smote}
Nitesh~V Chawla, Kevin~W Bowyer, Lawrence~O Hall, and W~Philip Kegelmeyer.
\newblock {SMOTE}: synthetic minority over-sampling technique.
\newblock {\em Journal of Artificial Intelligence Research (JAIR)},
  16:321--357, 2002.

\bibitem{feichtenhofer2018have}
Christoph Feichtenhofer, Axel Pinz, Richard~P Wildes, and Andrew Zisserman.
\newblock What have we learned from deep representations for action
  recognition?
\newblock In {\em Conference on Computer Vision and Pattern Recognition
  (CVPR)}, pages 7844--7853, 2018.

\bibitem{feldman2015certifying}
Michael Feldman, Sorelle~A Friedler, John Moeller, Carlos Scheidegger, and
  Suresh Venkatasubramanian.
\newblock Certifying and removing disparate impact.
\newblock In {\em ACM SIGKDD International Conference on Knowledge Discovery
  and Data Mining (KDD)}, pages 259--268, 2015.

\bibitem{fernando2013unsupervised}
Basura Fernando, Amaury Habrard, Marc Sebban, and Tinne Tuytelaars.
\newblock Unsupervised visual domain adaptation using subspace alignment.
\newblock In {\em International Conference on Computer Vision (ICCV)}, pages
  2960--2967, 2013.

\bibitem{girdhar2017actionvlad}
Rohit Girdhar, Deva Ramanan, Abhinav Gupta, Josef Sivic, and Bryan Russell.
\newblock Actionvlad: Learning spatio-temporal aggregation for action
  classification.
\newblock In {\em Conference on Computer Vision and Pattern Recognition
  (CVPR)}, pages 3165--3174, 2017.

\bibitem{gkioxari2015contextual}
Georgia Gkioxari, Ross Girshick, and Jitendra Malik.
\newblock Contextual action recognition with r*cnn.
\newblock In {\em International Conference on Computer Vision (ICCV)}, pages
  1080--1088, 2015.

\bibitem{goodfellow2014generative}
Ian Goodfellow, Jean Pouget-Abadie, Mehdi Mirza, Bing Xu, David Warde-Farley,
  Sherjil Ozair, Aaron Courville, and Yoshua Bengio.
\newblock Generative adversarial nets.
\newblock In {\em Advances in Neural Information Processing Systems (NIPS)},
  pages 2672--2680, 2014.

\bibitem{hardt2016equality}
Moritz Hardt, Eric Price, Nati Srebro, et~al.
\newblock Equality of opportunity in supervised learning.
\newblock In {\em Advances in Neural Information Processing Systems (NIPS)},
  pages 3315--3323, 2016.

\bibitem{he2016deep}
Kaiming He, Xiangyu Zhang, Shaoqing Ren, and Jian Sun.
\newblock Deep residual learning for image recognition.
\newblock In {\em Conference on Computer Vision and Pattern Recognition
  (CVPR)}, pages 770--778, 2016.

\bibitem{huang2018makes}
De-An Huang, Vignesh Ramanathan, Dhruv Mahajan, Lorenzo Torresani, Manohar
  Paluri, Li Fei-Fei, and Juan~Carlos Niebles.
\newblock What makes a video a video: Analyzing temporal information in video
  understanding models and datasets.
\newblock In {\em Conference on Computer Vision and Pattern Recognition
  (CVPR)}, pages 7366--7375, 2018.

\bibitem{jhuang2013towards}
Hueihan Jhuang, Juergen Gall, Silvia Zuffi, Cordelia Schmid, and Michael~J
  Black.
\newblock Towards understanding action recognition.
\newblock In {\em International Conference on Computer Vision (ICCV)}, pages
  3192--3199, 2013.

\bibitem{ji20133d}
Shuiwang Ji, Wei Xu, Ming Yang, and Kai Yu.
\newblock 3d convolutional neural networks for human action recognition.
\newblock {\em IEEE Transactions on Pattern Analysis and Machine Intelligence
  (PAMI)}, 35(1):221--231, 2013.

\bibitem{kay2017kinetics}
Will Kay, Joao Carreira, Karen Simonyan, Brian Zhang, Chloe Hillier, Sudheendra
  Vijayanarasimhan, Fabio Viola, Tim Green, Trevor Back, Paul Natsev, et~al.
\newblock The kinetics human action video dataset.
\newblock {\em arXiv preprint arXiv:1705.06950}, 2017.

\bibitem{khosla2012undoing}
Aditya Khosla, Tinghui Zhou, Tomasz Malisiewicz, Alexei~A Efros, and Antonio
  Torralba.
\newblock Undoing the damage of dataset bias.
\newblock In {\em European Conference on Computer Vision (ECCV)}, pages
  158--171, 2012.

\bibitem{kuehne2011hmdb}
Hildegard Kuehne, Hueihan Jhuang, Est{\'\i}baliz Garrote, Tomaso Poggio, and
  Thomas Serre.
\newblock {HMDB}: a large video database for human motion recognition.
\newblock In {\em International Conference on Computer Vision (ICCV)}, pages
  2556--2563, 2011.

\bibitem{laptev2005space}
Ivan Laptev.
\newblock On space-time interest points.
\newblock {\em International Journal of Computer Vision (IJCV)},
  64(2-3):107--123, 2005.

\bibitem{lecun1998gradient}
Yann LeCun, L{\'e}on Bottou, Yoshua Bengio, and Patrick Haffner.
\newblock Gradient-based learning applied to document recognition.
\newblock {\em Proceedings of the IEEE}, 86(11):2278--2324, 1998.

\bibitem{li2018resound}
Yingwei Li, Yi Li, and Nuno Vasconcelos.
\newblock {RESOUND}: Towards action recognition without representation bias.
\newblock In {\em European Conference on Computer Vision (ECCV)}, pages
  513--528, 2018.

\bibitem{mahdisoltani2018fine}
Farzaneh Mahdisoltani, Guillaume Berger, Waseem Gharbieh, David Fleet, and
  Roland Memisevic.
\newblock Fine-grained video classification and captioning.
\newblock {\em arXiv preprint arXiv:1804.09235}, 2018.

\bibitem{oquab2014learning}
Maxime Oquab, Leon Bottou, Ivan Laptev, and Josef Sivic.
\newblock Learning and transferring mid-level image representations using
  convolutional neural networks.
\newblock In {\em Conference on Computer Vision and Pattern Recognition
  (CVPR)}, pages 1717--1724, 2014.

\bibitem{ritter2017cognitive}
Samuel Ritter, David~GT Barrett, Adam Santoro, and Matt~M Botvinick.
\newblock Cognitive psychology for deep neural networks: A shape bias case
  study.
\newblock In {\em International Conference on Machine Learning (ICML)}, pages
  2940--2949, 2017.

\bibitem{russakovsky2015imagenet}
Olga Russakovsky, Jia Deng, Hao Su, Jonathan Krause, Sanjeev Satheesh, Sean Ma,
  Zhiheng Huang, Andrej Karpathy, Aditya Khosla, Michael Bernstein, et~al.
\newblock Imagenet large scale visual recognition challenge.
\newblock {\em International Journal of Computer Vision (IJCV)},
  115(3):211--252, 2015.

\bibitem{simonyan2014two}
Karen Simonyan and Andrew Zisserman.
\newblock Two-stream convolutional networks for action recognition in videos.
\newblock In {\em Advances in Neural Information Processing Systems (NIPS)},
  pages 568--576, 2014.

\bibitem{soomro2012ucf101}
Khurram Soomro, Amir~Roshan Zamir, and Mubarak Shah.
\newblock {UCF101}: A dataset of 101 human actions classes from videos in the
  wild.
\newblock {\em arXiv preprint arXiv:1212.0402}, 2012.

\bibitem{stock2018convnets}
Pierre Stock and Moustapha Cisse.
\newblock Convnets and imagenet beyond accuracy: Understanding mistakes and
  uncovering biases.
\newblock In {\em European Conference on Computer Vision (ECCV)}, pages
  498--512, 2018.

\bibitem{tommasi2017deeper}
Tatiana Tommasi, Novi Patricia, Barbara Caputo, and Tinne Tuytelaars.
\newblock A deeper look at dataset bias.
\newblock In {\em Domain Adaptation in Computer Vision Applications}, pages
  37--55. Springer, 2017.

\bibitem{torralba2003contextual}
Antonio Torralba.
\newblock Contextual priming for object detection.
\newblock {\em International Journal of Computer Vision (IJCV)},
  53(2):169--191, 2003.

\bibitem{torralba2011unbiased}
Antonio Torralba and Alexei~A Efros.
\newblock Unbiased look at dataset bias.
\newblock In {\em Conference on Computer Vision and Pattern Recognition
  (CVPR)}, pages 1521--1528, 2011.

\bibitem{tran2015learning}
Du Tran, Lubomir Bourdev, Rob Fergus, Lorenzo Torresani, and Manohar Paluri.
\newblock Learning spatiotemporal features with 3d convolutional networks.
\newblock In {\em International Conference on Computer Vision (ICCV)}, pages
  4489--4497, 2015.

\bibitem{wang2011action}
Heng Wang, Alexander Kl{\"a}ser, Cordelia Schmid, and Cheng-Lin Liu.
\newblock Action recognition by dense trajectories.
\newblock In {\em Conference on Computer Vision and Pattern Recognition
  (CVPR)}, pages 3169--3176, 2011.

\bibitem{wang2013action}
Heng Wang and Cordelia Schmid.
\newblock Action recognition with improved trajectories.
\newblock In {\em International Conference on Computer Vision (ICCV)}, pages
  3551--3558, 2013.

\bibitem{wang2016temporal}
Limin Wang, Yuanjun Xiong, Zhe Wang, Yu Qiao, Dahua Lin, Xiaoou Tang, and Luc
  Van~Gool.
\newblock Temporal segment networks: Towards good practices for deep action
  recognition.
\newblock In {\em European Conference on Computer Vision (ECCV)}, pages 20--36,
  2016.

\bibitem{yue2015beyond}
Joe Yue-Hei~Ng, Matthew Hausknecht, Sudheendra Vijayanarasimhan, Oriol Vinyals,
  Rajat Monga, and George Toderici.
\newblock Beyond short snippets: Deep networks for video classification.
\newblock In {\em Conference on Computer Vision and Pattern Recognition
  (CVPR)}, pages 4694--4702, 2015.

\bibitem{zemel2013learning}
Richard Zemel, Yu Wu, Kevin Swersky, Toni Pitassi, and Cynthia Dwork.
\newblock Learning fair representations.
\newblock In {\em International Conference on Machine Learning (ICML)}, pages
  325--333, 2013.

\bibitem{zhao2017men}
Jieyu Zhao, Tianlu Wang, Mark Yatskar, Vicente Ordonez, and Kai-Wei Chang.
\newblock Men also like shopping: Reducing gender bias amplification using
  corpus-level constraints.
\newblock In {\em Conference on Empirical Methods in Natural Language
  Processing (EMNLP)}, 2017.

\bibitem{zhou2018temporal}
Bolei Zhou, Alex Andonian, Aude Oliva, and Antonio Torralba.
\newblock Temporal relational reasoning in videos.
\newblock In {\em The European Conference on Computer Vision (ECCV)}, pages
  803--818, 2018.

\end{thebibliography}
	}
	
\end{document}